\documentclass[conference]{IEEEtran}

\usepackage{cite}
\usepackage{amsmath,amssymb,amsfonts}
\usepackage{algorithmic}
\usepackage{graphicx}
\usepackage{textcomp}
\usepackage{xcolor}
\usepackage{listings}
\usepackage{url}

\usepackage{booktabs}   
\usepackage{amsmath}    

\newcommand{\REGAL}{REGAL} 

\begin{document}

\title{REGAL: A Registry-Driven Architecture for Deterministic Grounding of Agentic AI in Enterprise Telemetry}

\author{\IEEEauthorblockN{Yuvraj Agrawal}
\IEEEauthorblockA{\textit{Adobe Inc.} \\}
}

\maketitle

\begin{abstract}

Enterprise engineering organizations produce high-volume, heterogeneous telemetry from version control systems, CI/CD pipelines, issue trackers, and observability platforms. Large Language Models (LLMs) enable new forms of agentic automation, but grounding such agents on private telemetry raises three practical challenges: limited model context, locally defined semantic concepts, and evolving metric interfaces.

We present REGAL, a registry-driven architecture for deterministic grounding of agentic AI systems in enterprise telemetry. REGAL adopts an explicitly architectural approach: deterministic telemetry computation is treated as a first-class primitive, and LLMs operate over a bounded, version-controlled action space rather than raw event streams.

The architecture combines (1) a Medallion ELT pipeline that produces replayable, semantically compressed Gold artifacts, and (2) a registry-driven compilation layer that synthesizes Model Context Protocol (MCP) tools from declarative metric definitions. The registry functions as an “interface-as-code” layer, ensuring alignment between tool specification and execution, mitigating tool drift, and embedding governance policies directly at the semantic boundary.

A prototype implementation and case study validate the feasibility of deterministic grounding and illustrate its implications for latency, token efficiency, and operational governance. This work systematizes an architectural pattern for enterprise LLM grounding; it does not propose new learning algorithms, but rather elevates deterministic computation and semantic compilation to first-class design primitives for agentic systems.
\end{abstract}


\section{Introduction}

Modern software engineering organizations generate distributed, high-volume telemetry across source control, CI/CD, issue trackers, and observability platforms. This telemetry is crucial for operational decision-making, but it is fragmented, schema-volatile, and often access-controlled. Naïvely exposing raw logs to probabilistic reasoners (e.g., LLMs via Retrieval-Augmented Generation) leads to three practical failures: (1) context overload and prohibitive token costs, (2) semantic ambiguity where organizational meanings are not globally defined, and (3) interface drift when hand-coded tools diverge from evolving telemetry semantics.

LLM-based agents promise to automate cross-system synthesis, but their usefulness depends on how we expose telemetry to them. In practice, we must decide whether the model should process raw events or interact with a semantically stable API surface. This paper advocates the latter: deterministic data computation should be performed explicitly and versioned; LLMs should be consumers of those deterministic artifacts and selectors of a bounded set of semantic tools.
We formalize this approach in \textbf{REGAL} (Registry-Driven Architecture for Grounded Agentic LLMs), the reference architecture presented in this paper. REGAL operationalizes deterministic grounding by compiling declarative metric definitions into a bounded semantic interface that LLM agents consume.

\subsection{Problem summary: limitations of probabilistic grounding}

Three constraints make grounding LLMs on enterprise telemetry challenging:

\begin{itemize}
  \item \textbf{Context limits.} Enterprise telemetry volumes far exceed contemporary model context sizes. Sending raw events to the model is economically and operationally infeasible for many tasks.
  \item \textbf{Local semantics.} Organizational concepts (``P1'', ``release-candidate'', ``regression'') are defined internally; probabilistic retrieval without deterministic binding leads to inconsistent interpretation and hallucination.
  \item \textbf{Interface evolution.} Hand-coded tools and ad-hoc APIs diverge from data semantics as schemas and metric definitions evolve, causing tool drift and governance risk.
\end{itemize}

These constraints suggest that grounding should be an architectural decision: compute, compress, and version telemetry first; expose a controlled action space second.

\subsection{Key idea: registry-driven semantic compilation}

We introduce a design in which a declarative \emph{metrics registry} is the single source of truth. The registry captures metric semantics, retrieval logic, and governance metadata; a deterministic compilation step generates the concrete MCP-exposed tool surface from this registry. In effect, the registry implements an “interface-as-code” pattern for agentic systems: rather than hand-writing tools, operators declare metrics and compilation produces executable, audited, and access-controlled tools.

This pattern directly addresses tool drift: tool specifications and their implementations are derived from the same artifact, ensuring alignment over time. It also constrains the agent’s action space, which materially reduces the hallucination surface and simplifies policy enforcement.

\subsection{Scope and contributions}

REGAL is presented as a systematization and reference architecture rather than an algorithmic contribution. The core contributions are:

\begin{itemize}
  \item \textbf{Registry-driven MCP compilation:} An interface-as-code pattern that synthesizes tool schemas, retrieval logic, and governance controls from declarative metric definitions.
  \item \textbf{Deterministic grounding architecture:} A practical reference architecture that separates deterministic telemetry computation (Medallion ELT) from probabilistic reasoning (LLM agents).
  \item \textbf{Prototype validation and lessons learned:} A prototype implementation and case study demonstrating feasibility and describing operational trade-offs (latency, token scaling, governance).
\end{itemize}

\vspace{4pt}
\noindent\textbf{Novelty statement} To the best of our knowledge, \emph{REGAL} is the first systematization that (a) formalizes registry-driven semantic compilation as an explicit architectural grounding constraint for enterprise LLM agents, and (b) operationalizes that constraint end-to-end—covering deterministic ingestion, versioned Medallion transformations, and automated compilation of declarative metric definitions into MCP-exposed, access-controlled tools. While elements of the pipeline (e.g., Medallion ELT, change streams, MCP tooling) exist independently in prior work, REGAL's novel combination is the explicit architectural commitment to treat the registry as the single source of truth and the compilation step as the enforcement mechanism that bounds the agent's action space.
\vspace{4pt}

\begin{figure}[htbp]
  \centering
  \includegraphics[width=0.9\columnwidth]{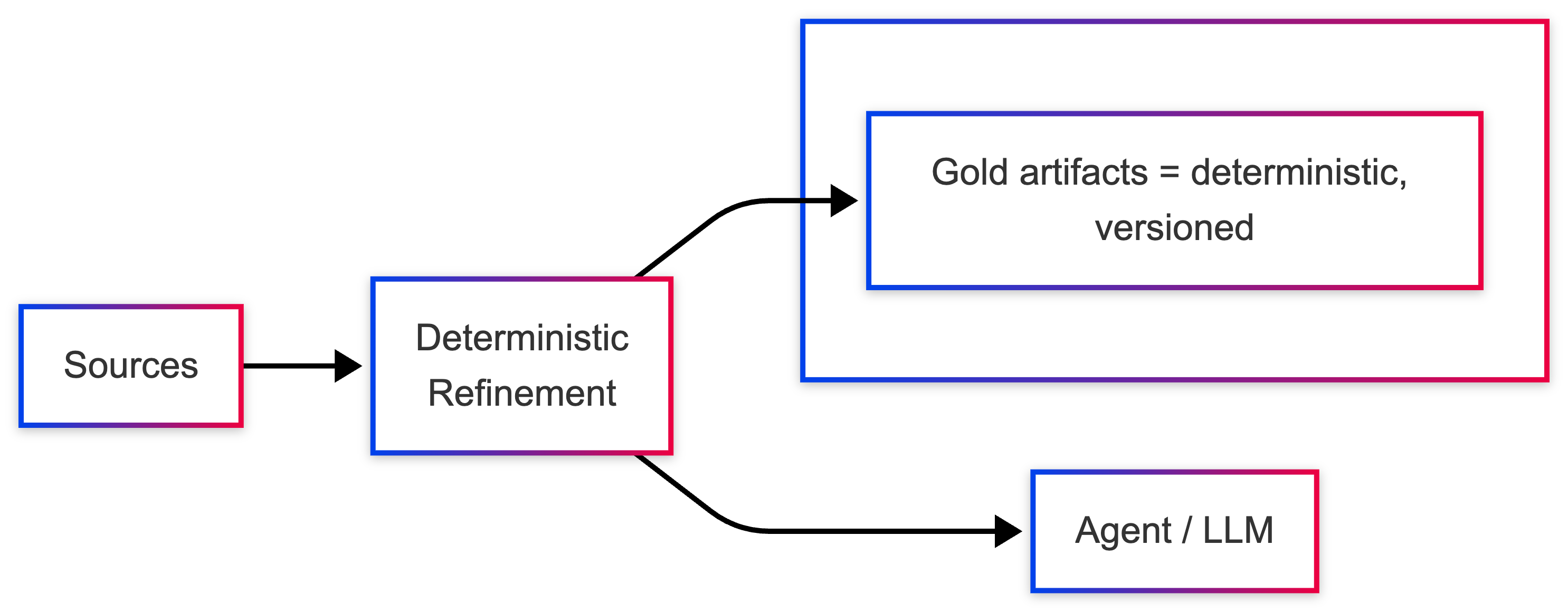}
  \caption{High-level architecture: sources are deterministically refined into versioned Gold artifacts; agentic LLMs consume only those artifacts. The thick boundary emphasizes the deterministic \(\rightarrow\) probabilistic handoff.}
  \label{fig:overview}
\end{figure}

Figure~\ref{fig:overview} provides a visual summary of the architecture: deterministic refinement produces versioned Gold artifacts that are the sole inputs to LLM-based agents.

\subsection{Paper roadmap}

The remainder of the paper proceeds as follows. Section II presents the reference architecture and the deterministic/probabilistic separation. Section III details the registry-driven semantic compilation pattern (the paper’s primary contribution). Section IV describes the deterministic ingestion and Medallion pipeline that produces Gold artifacts. Section V discusses storage and serving considerations; Section VI describes the push/pull interaction model. Section VII positions this approach relative to RAG, Text-to-SQL and vendor platforms. Section VIII presents the prototype implementation and case study. We conclude with open research directions in Section IX.


\section{The Reference Architecture}

Designing reliable agentic AI for engineering observability requires more than attaching an LLM to existing APIs. Enterprise telemetry is heterogeneous, temporal, and governed by access controls; it is also subject to rate limits, schema drift, and transient failures. Exposing raw telemetry to probabilistic reasoners therefore risks costly token usage, ambiguous semantics, and brittle tooling.

The \REGAL{} architecture centers a simple design principle:
 perform deterministic computation and semantic compilation before exposing any telemetry to probabilistic agents. In practice, this means (a) produce replayable, version-controlled Gold artifacts via a deterministic ELT pipeline, and (b) synthesize a bounded, audited action space (tools) from a declarative metrics registry. The registry-to-tool compilation pattern is the primary architectural contribution of this work: it is the mechanism by which the system constrains reasoning to a safe, auditable interface.

\subsection{Architectural overview}

The system separates responsibilities into four layers with a unidirectional data flow:

\begin{enumerate}
  \item \textbf{Source layer.} External systems (version control, CI/CD, issue trackers, observability) that emit events or state.
  \item \textbf{Ingestion \& orchestration (write path).} Deterministic extraction, validation, reconciliation, and upsert semantics that guarantee replayability.
  \item \textbf{Medallion storage (context store).} A layered transformation that converts raw telemetry (Bronze) to harmonized records (Silver) and to compact, semantically rich Gold artifacts intended for AI consumption.
  \item \textbf{Semantic layer (read path).} A registry-driven compilation component that generates concrete MCP-exposed tools, access controls, and caching policies; agents reason only over these synthesized tools and their deterministic outputs.
\end{enumerate}

\subsection{Deterministic–probabilistic non-interference (concise property)}

To make the separation explicit, we state a non-interference property in plain terms:

\begin{quote}
Deterministic computation (ingestion, transformation, compilation) produces version-controlled artifacts and tool definitions. Probabilistic reasoning (LLM inference) may consume these artifacts, but probabilistic inference must not alter or redefine telemetry computation. In other words, interpretation is allowed to depend on computed artifacts, but computation must not depend on interpretation.
\end{quote}

\vspace{4pt}
\noindent\textbf{Formal property:}~
For clarity we adopt symbols:

\(\mathcal{D}\) — deterministic transformations (ingestion, harmonization, compilation),  
\(\mathcal{P}\) — probabilistic inference (LLM/agent reasoning),  
\(G\) — Gold artifacts, i.e., version-controlled outputs of \(\mathcal{D}\).

The intended information flow is
\[
\mathcal{D}(\text{input}) \;\longrightarrow\; G
\qquad\text{and}\qquad
\mathcal{P}(G) \;\longrightarrow\; \text{Output}.
\]

We express non-interference informally as
\[
\frac{\partial \mathcal{D}}{\partial \mathcal{P}} \;=\; 0,
\]
meaning that changes in \(\mathcal{P}\) (model choice, prompt design, or reasoning strategy) must not alter \(\mathcal{D}\) or \(G\). Operationally, \(\mathcal{D}\) must be replayable, versioned, and invariant under substitution of \(\mathcal{P}\).
\vspace{4pt}

This non-interference property is enforced operationally by version-controlled transformations, deterministic upsert semantics, and compilation-from-registry (so tool definition and implementation originate from a single artifact).

\subsection{Hybrid push–pull interaction model}

Practical agentic systems must support both retrospective analysis and near-real-time awareness:

\begin{itemize}
  \item \textbf{Pull (historical reasoning).} On user request, agents invoke registry-generated tools that return deterministic Gold artifacts covering bounded time windows. Because the artifacts are pre-aggregated, agents avoid token-expensive reconstruction of raw logs.
  \item \textbf{Push (event-driven awareness).} Change streams or event monitors propagate state transitions in Gold artifacts to alerting and agent triggers. This decouples detection latency from user invocation frequency and supports proactive workflows.
\end{itemize}

Both paths operate over the same Gold artifacts and compiled tool surface, eliminating semantic divergence between reactive and proactive workflows.

\subsection{Design principles and invariants (operational form)}

Rather than presenting ornamental equations, we express the architecture’s key invariants as operational guarantees:

\paragraph{Model-agnostic data invariant.} Transformations that produce Gold artifacts are independent of downstream inference models. Swapping the LLM or agent framework does not require changing ingestion, harmonization, or metric computation.

\paragraph{Idempotent ingestion and replayability.} Ingestion and transformation are implemented so that repeated execution over the same input yields identical Gold artifacts. This is achieved via deterministic keys, upsert semantics, and replayable DAGs.

\paragraph{Context compression as semantic engineering.} The Medallion stages progressively increase semantic density and decrease raw volume so that the Gold working set is compact, semantically explicit, and suitable for bounded LLM consumption.

\paragraph{Semantic consistency via single-source registry.} Metric semantics, retrieval logic, and governance metadata are declared in a single registry artifact. Tool schemas, implementations, access controls, and caching policies are compiled from this artifact, ensuring consistent definitions across runtime components.

\subsection{Operational consequences}

Framing grounding as an architectural constraint produces several practical benefits:

\begin{itemize}
  \item \textbf{Auditability:} All metric definitions and transformations are version-controlled and replayable.
  \item \textbf{Governance:} Access policies and TTLs are enforced at the compiled tool boundary, simplifying compliance.
  \item \textbf{Reduced hallucination surface:} The agent’s action space is bounded to a finite set of compiled tools, reducing opportunities for unconstrained model synthesis.
  \item \textbf{Predictable economics:} Upstream aggregation bounds token growth by exposing compact artifacts instead of raw event streams.
\end{itemize}

The next sections make the registry-driven compilation pattern concrete, describe the deterministic ingestion pipeline that produces Gold artifacts, and discuss design trade-offs observed in our prototype.


\section{The Semantic Layer: Registry-Driven Semantic Compilation}

A consistent storage layer is necessary but insufficient for reliable agentic reasoning. Large Language Models (LLMs) do not natively understand organizational schema semantics, nor do they enforce alignment between a tool’s declared purpose and its underlying implementation. Bridging structured telemetry and probabilistic reasoning therefore requires an explicit semantic interface that constrains how telemetry is exposed.

The core contribution of this work is a registry-driven semantic compilation layer. Instead of manually defining tool schemas and API handlers independently—a process prone to divergence over time—we define metrics declaratively in a centralized registry and compile the concrete Model Context Protocol (MCP) tool surface from that registry. The registry acts as a single source of truth for metric meaning, retrieval logic, and governance metadata.

\begin{figure}[htbp]
  \centering
  \includegraphics[width=\columnwidth]{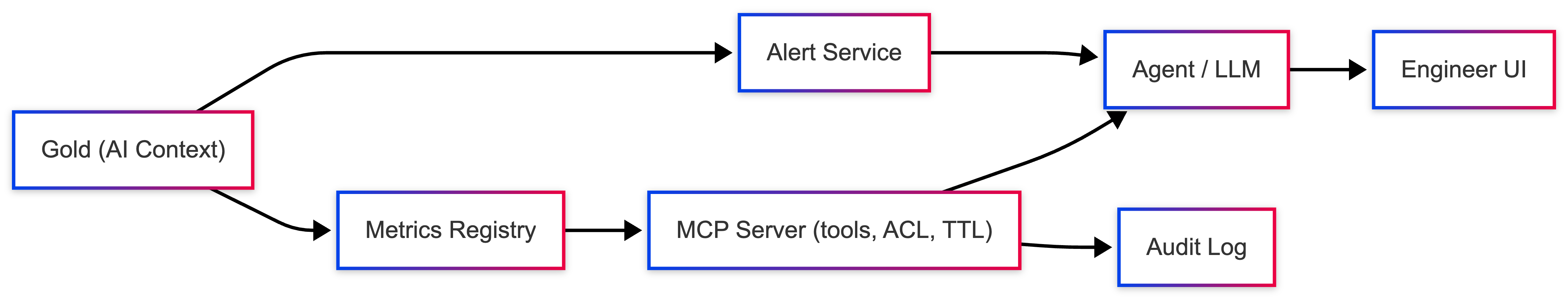}
  \caption{Read path and serving: the registry compiles declarative metric definitions into MCP tools; agents invoke MCP tools (pull) and the Gold layer can also drive alerts (push). Governance (ACLs, audit) is enforced at the MCP boundary.}
  \label{fig:readpath}
\end{figure}

Figure~\ref{fig:readpath} illustrates the read path: deterministic Gold artifacts are exposed through a compiled MCP interface, and agents interact only with this bounded tool surface.

\subsection{Tool Drift as a Systems Failure Mode}

In many LLM-integrated systems, the tool definition visible to the model and the backend implementation evolve independently. Over time, the model’s understanding of what a tool does may diverge from the actual computation executed by the system. We refer to this misalignment as \emph{tool drift}.

In enterprise settings, tool drift is not merely an inconvenience. It creates governance risk: metric definitions may change without corresponding updates to prompts or tool documentation, leading to incorrect interpretations that are difficult to audit.

The registry-driven approach addresses this by construction: tool schemas, descriptions, and retrieval logic are generated from a single declarative artifact. There is no manually maintained duplication between interface and implementation.

\subsection{The Registry as an Interface Definition Layer}

The metrics registry captures, for each metric:

\begin{itemize}
    \item A stable identifier and human-readable description,
    \item The deterministic retrieval function over Gold artifacts,
    \item Platform or environment scoping,
    \item Governance metadata (e.g., caching policy, access control category).
\end{itemize}

At initialization time, a compilation step generates:

\begin{itemize}
    \item Concrete MCP tool schemas,
    \item Tool descriptions presented to the LLM,
    \item Access control bindings,
    \item Caching behavior derived from metric volatility.
\end{itemize}

This pattern can be understood as “interface-as-code”: instead of hand-writing tool endpoints, engineers declare metric semantics and allow compilation to produce a consistent runtime interface.

\noindent\textbf{Implementation independence} The registry-to-tool compilation pattern is language- and framework-agnostic. While prototype examples and snippets in this paper use Python for clarity, the registry can target any interface description (OpenAPI/JSON Schema, gRPC proto, or other platform-specific bindings). The key property is that the operational contract (semantic definition, retrieval logic, ACLs, and caching policies) originates from the registry; the compilation backend may emit language-specific adapters as required by the deployment environment.

\subsection{Bounded Action Space and Semantic Closure}

Because tools are compiled from a finite registry, the agent’s action space is explicitly bounded. Agents cannot invoke arbitrary queries or synthesize new metrics at runtime; they must select from the compiled tool set.

This bounded action space has two practical consequences:

\begin{itemize}
    \item \textbf{Reduced hallucination surface.} The LLM selects among predefined computational primitives rather than generating unconstrained queries.
    \item \textbf{Governance enforcement at the boundary.} Access control and caching policies are applied uniformly at the tool interface.
\end{itemize}

In this design, the LLM does not reason over raw telemetry. It reasons over deterministic outputs returned by well-defined metric tools.

\subsection{Why Not Text-to-SQL?}

A natural alternative is to expose the Gold schema directly and allow the LLM to generate SQL queries.

While expressive, this approach introduces two risks in enterprise environments:

\begin{itemize}
    \item \textbf{Unbounded query surface.} SQL generation exposes the full schema and join space to probabilistic decoding. Erroneous joins, unbounded scans, or semantically incorrect filters can produce misleading results.
    \item \textbf{Governance and safety complexity.} Enforcing read-only guarantees, cost limits, and semantic constraints requires additional sandboxing and validation layers.
\end{itemize}

The registry-driven MCP approach constrains the LLM to a finite, version-controlled semantic API. Rather than synthesizing arbitrary queries, the model selects from curated metric tools whose behavior is deterministic and auditable. This reduces operational risk and simplifies compliance.

\subsection{Caching and Access Control as Interface Policies}

Caching and access control are treated as interface-level policies rather than database implementation details.

Metrics are categorized by volatility, and caching durations are derived from that categorization. Similarly, user identity and permission scopes are propagated to the MCP layer, where tool invocation is authorized or denied based on declared metric categories.

By embedding these policies into the compilation process, governance becomes a property of the interface itself rather than an afterthought layered onto raw query access.

\subsection*{Summary}

The registry-driven semantic compilation layer is the central architectural mechanism that enables deterministic grounding. It ensures alignment between tool specification and execution, bounds the agent’s action space, and embeds governance directly into the runtime interface. The next section describes how deterministic ingestion and Medallion transformations produce the Gold artifacts on which this semantic layer depends.


\section{The Data Foundation: Deterministic Ingestion and Transformation}

The reliability of an agentic AI system is bounded by the reliability of its data pipeline. If telemetry ingestion is non-deterministic, inconsistent under retry, or sensitive to transient failures, then downstream reasoning inherits those inconsistencies.

Accordingly, the Write Path is designed around three operational guarantees:

\begin{itemize}
    \item \textbf{Replayability:} Re-running ingestion over the same input produces the same Gold artifacts.
    \item \textbf{Idempotency:} Retries do not duplicate or corrupt metric state.
    \item \textbf{Versioned transformation:} Metric definitions evolve explicitly and historical recomputation is supported.
\end{itemize}

\begin{figure}[htbp]
  \centering
  \includegraphics[width=\columnwidth]{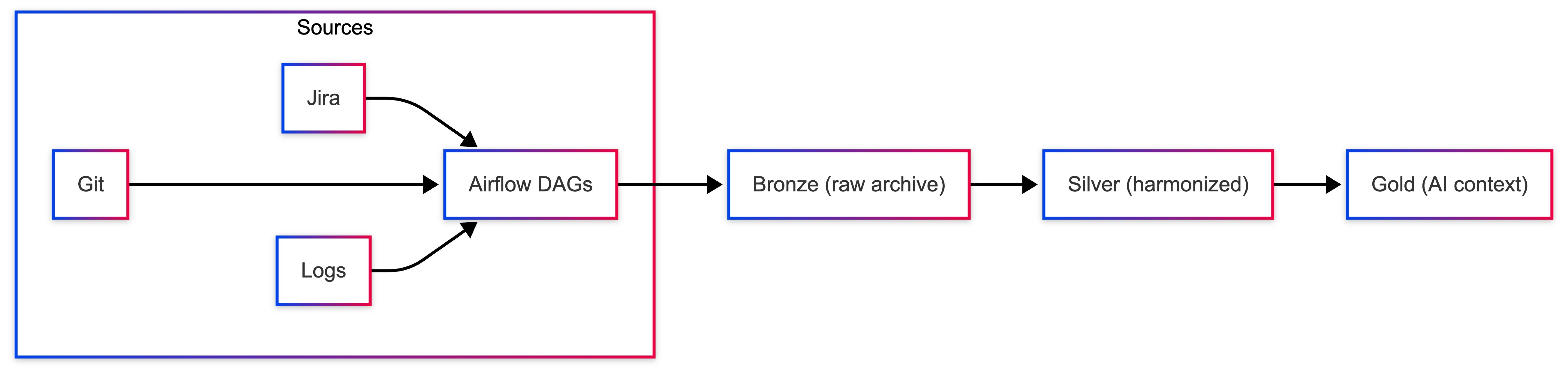}
  \caption{Write path (Medallion ELT): ingestion, deterministic upserts, Bronze (raw archive), Silver (schema harmonization), and Gold (AI context artifacts).}
  \label{fig:writepath}
\end{figure}

Figure~\ref{fig:writepath} illustrates the transformation pipeline from heterogeneous sources to semantically compressed Gold artifacts.

\subsection{Source Heterogeneity and Extraction Patterns}

Enterprise telemetry originates from systems with differing consistency and mutation models. In practice, we observe three recurring ingestion patterns:

\paragraph{State-based APIs.}
Systems such as issue trackers and code repositories expose mutable entity state. Incremental extraction relies on monotonic update markers (e.g., timestamps or cursors). Re-execution over the same time window converges to the same logical state.

\paragraph{Event-based logs.}
Observability systems emit append-only event streams. Rather than forwarding raw logs to the agent, we perform windowed aggregation at ingestion time, converting high-entropy event streams into compact metric summaries.

\paragraph{Snapshot datasets.}
Some systems publish periodic full-state snapshots. These are ingested using replace-window semantics: the snapshot defines authoritative state for that interval.

This taxonomy is not novel; it reflects established backend practice. Its relevance here lies in making deterministic ingestion explicit as a prerequisite for reliable LLM grounding.

\subsection{Idempotency and Convergence in Practice}

Idempotency is implemented through deterministic record identifiers and upsert semantics. Each metric record is associated with a stable identity derived from source attributes (e.g., source identifier, timestamp, platform, metric name). Retries update existing records rather than inserting duplicates.

Operationally, this ensures:

\begin{itemize}
    \item Replaying a failed job does not change logical Gold state.
    \item Concurrent ingestion tasks converge to a single consistent representation.
    \item Agents never observe duplicated or partially applied metrics.
\end{itemize}

Rather than presenting idempotency as a mathematical novelty, we treat it as a necessary systems guarantee that underpins deterministic grounding.

\subsection{Execution Model and Throughput Considerations}

Ingestion workloads are typically I/O-bound. We use a task orchestration engine (e.g., Airflow) to coordinate extraction, transformation, and storage steps, and apply parallelism at both the task and intra-task levels.

The specific concurrency primitives are not central to the architecture; the important property is that execution order and retries do not affect final Gold artifacts. Deterministic semantics are preserved regardless of scheduling interleavings.

(Implementation details such as thread pools and worker counts are omitted here for brevity; they do not alter the architectural guarantees.)

\subsection{Medallion Transformation as Deterministic Refinement}

We adopt a Medallion-style refinement pipeline:

\begin{itemize}
    \item \textbf{Bronze:} Immutable raw payload archive.
    \item \textbf{Silver:} Schema-harmonized and validated records.
    \item \textbf{Gold:} Deterministically computed metric artifacts intended for agent consumption.
\end{itemize}

The role of this pipeline is not merely data hygiene. It serves two architectural purposes:

\begin{enumerate}
    \item To make metric computation explicit and version-controlled.
    \item To reduce entropy before LLM interaction, exposing compact, semantically meaningful artifacts.
\end{enumerate}

For example, multiple source fields (e.g., merged timestamps or resolution timestamps) may be normalized into a unified event timestamp during Silver-stage harmonization. Such transformations are declarative and versioned.

\subsection{Versioned Transformation and Backfill}

Metric definitions evolve. To preserve longitudinal consistency, transformations are versioned. When a definition changes, historical Gold artifacts can be recomputed from immutable Bronze data.

This supports:

\begin{itemize}
    \item Reproducible historical analysis,
    \item Explicit evolution of metric semantics,
    \item Alignment between registry definitions and stored artifacts.
\end{itemize}

Deterministic ingestion and versioned transformation ensure that the semantic layer described in the previous section operates over stable, replayable computational artifacts rather than transient event streams.

\noindent\textbf{Implementation note:} The prototype implements a tiered-parallel fetcher (inter-task parallelism via the orchestration engine and intra-task concurrency via thread pools). A concise, illustrative Python listing is provided in Appendix~\ref{app:impl} for reference; the specific concurrency primitives (thread vs. process pools, worker counts) are deployment-dependent and omitted here for clarity.


\section{Storage Design and Serving Constraints}

The storage layer determines whether deterministic artifacts can be retrieved within interactive latency bounds. For agentic systems, overall response time consists of (a) artifact retrieval and (b) model inference. The architectural objective is to ensure that artifact retrieval remains predictably fast and does not dominate end-to-end interaction time.

Rather than focusing on database-specific optimizations, we align physical storage layout with observed access patterns and the semantic structure of Gold artifacts.

\subsection{Workload-Aligned Storage}

Gold artifacts are typically accessed via bounded time-range queries scoped by platform and metric. The dominant pattern resembles:

\[
\mathrm{Query}(\text{metric}, \text{platform}, t_1:t_2).
\]

To support this pattern efficiently, storage is organized using time-bucketed collections with compact metadata keys. This reduces per-record overhead, improves locality for time-range scans, and supports predictable query latency under moderate concurrency.

The goal is not maximal throughput, but bounded and stable latency for common metric retrieval operations.

\subsection{Layer-Specific Granularity}

Different stages of the Medallion pipeline have different workload characteristics:

\begin{itemize}
    \item \textbf{Bronze and Silver layers} prioritize ingestion fidelity and write throughput.
    \item \textbf{Gold layer} prioritizes read efficiency for aggregated metrics.
\end{itemize}

Accordingly, finer-grained storage is used for raw and harmonized data, while coarser-grained organization is used for aggregated Gold artifacts. This reflects the architectural distinction between raw telemetry and AI-facing context artifacts.

\subsection{Indexing for Agent Access Patterns}

Unlike traditional dashboard usage, agentic systems may issue repeated or multi-dimensional metric queries during iterative reasoning. To support this safely and predictably, compound indices are defined over the dominant filter axes (e.g., platform, metric, timestamp).

This indexing strategy is not novel; it is standard database practice. Its relevance here lies in preserving the architectural invariant that retrieval latency remains bounded and subordinate to model inference time under representative workloads.

\subsection{Prototype Observations}

In prototype deployments where Gold artifacts fit in memory, aggregated metric retrieval remained consistently sub-second under moderate concurrent load, and cached responses were effectively immediate. In these scenarios, model inference latency dominated overall interaction time.

These observations validate that deterministic artifact retrieval can remain predictable and does not undermine the architectural separation between data computation and probabilistic reasoning.

\subsection*{Role Within the Architecture}

Storage is therefore not presented as a novel database contribution, but as an enabling layer: it ensures that deterministically computed Gold artifacts can be served within interactive constraints. The semantic compilation layer described earlier depends on this stability but does not rely on any database-specific mechanism beyond standard indexing and time-range optimization.


\section{Real-Time Intelligence: Event-Driven Push Path}

While the pull interface enables bounded historical reasoning, operational environments also require timely awareness of state transitions. A purely query-driven system delays anomaly detection until a user explicitly invokes a tool. The push path extends deterministic grounding into the temporal domain by propagating updates from Gold artifacts to monitoring and agent workflows.

\subsection{Streaming Rather Than Polling}

Traditional polling-based alert systems introduce latency proportional to the polling interval and may miss transient state changes between checks. In contrast, change-stream mechanisms emit events whenever Gold artifacts are inserted or updated.

Because these events are derived directly from deterministically computed Gold artifacts, push-based detection operates over semantically stable metrics rather than raw telemetry. This preserves the architectural separation between deterministic computation and probabilistic reasoning even in real-time workflows.

\subsection{Delivery Semantics and Operational Convergence}

Change streams provide at-least-once delivery semantics. In practice, this means that a given update may be delivered more than once. The alert handling logic is therefore implemented to be idempotent: processing the same update multiple times does not alter the final alert state.

Operationally, this ensures:

\begin{itemize}
    \item No missed state transitions under restart or transient failure,
    \item No duplicate notifications due to replayed events,
    \item Convergent alert state across restarts.
\end{itemize}

Rather than presenting convergence as a formal property, we treat it as a practical requirement for reliable event-driven systems.

\subsection{Relative Severity and Stability Controls}

Metrics vary in scale and interpretation across domains (e.g., crash rate versus deployment frequency). Severity is therefore evaluated relative to configured thresholds, allowing consistent escalation behavior across heterogeneous metric types.

Event-driven systems are also susceptible to oscillation near thresholds. To mitigate this, two stabilization mechanisms are applied:

\begin{itemize}
    \item \textbf{Hysteresis:} Resolution requires exceeding a safety margin beyond the threshold, preventing rapid flapping.
    \item \textbf{Cooldown windows:} Repeated alerts for the same metric and severity are rate-limited within a defined interval.
\end{itemize}

These mechanisms ensure that push-based detection remains operationally stable without altering underlying metric semantics.

\subsection{Shared Semantics Between Push and Pull}

A key architectural property is that both pull-based reasoning and push-based alerting operate over the same Gold artifacts and compiled metric definitions. There is no separate computation path for alerting.

This symmetry eliminates a common failure mode in observability systems, where dashboards and alert pipelines compute metrics differently. By reusing deterministic Gold artifacts across both interaction modes, the system maintains semantic consistency across retrospective analysis and real-time detection.


\section{Comparison to Existing Approaches}

Enterprise AI-assisted observability systems generally adopt one of four grounding strategies:

\begin{enumerate}
    \item Retrieval-Augmented Generation (RAG) over raw logs,
    \item Text-to-SQL over structured schemas,
    \item Dashboard-based LLM overlays,
    \item Vendor-integrated AI assistants tightly coupled to proprietary platforms.
\end{enumerate}

These approaches represent different design choices along four dimensions:

\begin{itemize}
    \item \textbf{Determinism:} Is metric computation definitionally fixed or probabilistically inferred?
    \item \textbf{Token Scaling:} Does context size scale with raw telemetry or with aggregated metrics?
    \item \textbf{Interaction Model:} Is the system purely pull-based, or does it support push-based propagation?
    \item \textbf{Semantic Governance:} Are metric definitions explicit, version-controlled, and auditable?
\end{itemize}

\begin{table}[htbp]
\caption{Architectural Comparison of AI Grounding Strategies}
\centering
\small
\begin{tabular}{p{1.4cm} p{1.4cm} p{1.4cm} p{1.3cm} p{1.4cm}}
\toprule
\textbf{Architecture} & \textbf{Determinism} & \textbf{Token Scaling} & \textbf{Push} & \textbf{Governance} \\
\midrule
RAG over Logs & Probabilistic & $\propto$ Retrieved Logs & No & Partial \\
Dashboard + LLM & Schema-Inferred & Aggregation-Dependent & Pull Only & Limited \\
Vendor AI Platforms & Platform-Defined & Aggregation-Dependent & Often & Vendor-Controlled \\
\textbf{Registry-Driven MCP} & Definitionally Fixed & $\propto$ Metric Set & Push + Pull & Registry-Controlled \\
\bottomrule
\end{tabular}
\label{tab:comparison}
\end{table}

Table~\ref{tab:comparison} summarizes these structural distinctions.

\subsection{Retrieval-Augmented Generation over Logs}

RAG systems retrieve top-$k$ log fragments using vector similarity and present them to the LLM for reasoning \cite{b8}. This approach is well-suited for unstructured knowledge tasks.

However, in enterprise telemetry contexts:

\begin{itemize}
    \item Token usage scales with the number of retrieved log entries.
    \item Retrieval is probabilistic and may surface semantically adjacent but operationally distinct events.
    \item Metric computation is implicitly performed inside the model rather than through deterministic aggregation.
\end{itemize}

The registry-driven architecture differs by performing metric aggregation upstream and exposing only deterministic Gold artifacts. This reduces token growth and ensures repeated tool invocation over identical state yields identical outputs.

\subsection{Why Not Text-to-SQL?}
Text-to-SQL systems translate natural language into executable SQL and can provide flexible access to structured datasets. However, in enterprise agentic workflows they introduce operational risks that are not addressed by syntactic correctness alone:

\begin{enumerate}
  \item \textbf{Execution safety:} Generated SQL can be syntactically valid but semantically harmful (e.g., unbounded scans, cross-tenant joins, or inadvertent exposure of sensitive fields) unless a robust sandbox and query-validation layer is present.
  \item \textbf{Resource and latency risk:} LLM-generated queries may lack appropriate predicates and cause expensive full-table scans or unpredictable execution times, undermining interactive responsiveness.
  \item \textbf{Governance complexity:} Allowing unconstrained query generation expands the agent's action surface, complicating audit trails, access control, and reproducibility.
\end{enumerate}

The registry-driven MCP approach instead exposes a bounded, version-controlled set of semantic primitives (tools) that map to deterministic Gold artifacts. This reduces the operational surface area: agents select audited computational primitives rather than generating arbitrary queries, simplifying execution safety, cost predictability, and governance.

\subsection{Dashboard-Based LLM Overlays}

Business intelligence platforms increasingly offer natural-language interfaces layered over pre-existing dashboards.

In these systems:

\begin{itemize}
    \item Column semantics are inferred from schema names and metadata.
    \item Interaction is typically pull-oriented.
    \item Tool definitions are not necessarily version-controlled alongside transformation logic.
\end{itemize}

While useful for exploratory analysis, such overlays may not guarantee alignment between metric definition and implementation as schemas evolve.

\subsection{Vendor-Integrated AI Assistants}

Observability vendors provide AI-driven anomaly detection and summarization within their proprietary ecosystems.

These systems often:

\begin{itemize}
    \item Encapsulate metric transformations within internal pipelines,
    \item Restrict cross-platform correlation to telemetry stored within a single vendor stack,
    \item Abstract away transformation logic from user control.
\end{itemize}

The registry-driven architecture differs by making metric definitions explicit and version-controlled, and by integrating heterogeneous sources into a unified Gold representation before inference.

\subsection{Architectural Trade-Offs}

The distinguishing property of the proposed architecture is not a new learning algorithm, but an explicit architectural constraint: deterministic computation and semantic compilation precede probabilistic reasoning.

Compared to RAG, the system reduces probabilistic retrieval from raw logs. Compared to Text-to-SQL, it bounds the agent’s action space. Compared to dashboard overlays and vendor assistants, it centralizes metric semantics in a version-controlled registry.

These trade-offs prioritize determinism, governance, and reproducibility over maximal query expressiveness. In regulated or operationally sensitive environments, such constraints may be preferable to fully generative query access.


\section{System Implementation and Architectural Case Study}

This section presents a prototype implementation and case study intended to validate the architectural feasibility of deterministic grounding. The objective is not to establish large-scale benchmark superiority, but to examine whether the proposed separation between deterministic computation and probabilistic reasoning holds under representative enterprise workloads.

The prototype integrates multiple telemetry sources and was deployed on commodity server hardware using standard data engineering components. The focus of evaluation is threefold:

\begin{enumerate}
    \item Reduction of cross-system investigation overhead,
    \item Preservation of interactive latency under agent usage,
    \item Practical impact of upstream semantic aggregation on token usage.
\end{enumerate}

\subsection{Case Study: Incident Investigation Workflow}

A central motivation for deterministic grounding is reducing investigation friction during operational incidents. We evaluated a representative scenario:

\textit{``Why did the iOS crash rate spike yesterday?''}

\subsubsection*{Manual Workflow}

In traditional workflows, engineers:

\begin{enumerate}
    \item Filter observability dashboards by platform and time window,
    \item Inspect recent CI/CD deployments,
    \item Cross-reference issue tracker tickets and change logs,
    \item Correlate findings across multiple interfaces.
\end{enumerate}

This process requires manual context switching and implicit reasoning across heterogeneous systems.

\subsubsection*{Agent-Assisted Workflow}

Under the proposed architecture:

\begin{enumerate}
    \item The engineer issues a natural-language query.
    \item The agent selects deterministic metric tools (e.g., stability, recent deployments).
    \item The agent synthesizes an explanation over returned Gold artifacts.
\end{enumerate}

\paragraph{Observational Outcome.}

Across representative scenarios, agent-assisted workflows reduced multi-step cross-system navigation into a single bounded interaction. While not statistically powered as a controlled user study, the case study demonstrates that deterministic compilation materially simplifies investigative reasoning by consolidating metric semantics into a single interface.

\subsection{Interactive Latency Characteristics}

For agentic systems to remain usable, artifact retrieval must not dominate overall response time.

Under moderate concurrent load in the prototype deployment:

\begin{itemize}
    \item Aggregated metric retrieval completed within interactive bounds,
    \item Cached queries were effectively immediate,
    \item End-to-end latency was dominated by model inference rather than data access.
\end{itemize}

These observations validate the architectural goal that deterministic retrieval remains subordinate to inference latency under typical workloads.

\subsection{Token Scaling Behavior}

Enterprise telemetry can span thousands of raw log events for a single investigative window. When grounding directly on raw logs, token consumption grows with the number of retrieved events.

In contrast, the registry-driven architecture exposes compact, pre-aggregated Gold artifacts. Token usage therefore scales with the number of selected metrics rather than the number of underlying log entries.

Prototype simulations over multi-hour telemetry windows confirmed that raw-log serialization produces token counts substantially larger than aggregated metric representations. While exact savings depend on workload characteristics, the qualitative reduction is significant and consistent with the architectural design.

\subsection{Limitations}

This implementation has several limitations:

\begin{itemize}
    \item \textbf{Prototype scope.} Results reflect a prototype deployment and representative workloads rather than controlled, large-scale benchmarks.
    \item \textbf{Limited sample size.} Case studies involve a small number of engineers and scenarios.
    \item \textbf{Model variability.} Latency characteristics depend on contemporary LLM inference performance.
    \item \textbf{Comparative baselines.} The system was not evaluated against a full Text-to-SQL or GraphRAG implementation under identical conditions.
\end{itemize}

The purpose of this section is therefore architectural validation rather than algorithmic benchmarking. The findings demonstrate feasibility and operational coherence of deterministic grounding, but do not claim statistical dominance over alternative grounding techniques.


\section{Conclusion and Future Directions}

This paper presented REGAL, a registry-driven architecture for grounding agentic AI systems in enterprise telemetry through deterministic data computation and semantic compilation. Rather than treating LLMs as primary processors of raw telemetry, the proposed design constrains probabilistic reasoning to operate exclusively over replayable, version-controlled artifacts.

The central contribution is the registry-driven compilation pattern: metric semantics, retrieval logic, and governance metadata are declared once and compiled into a bounded MCP tool interface. This “interface-as-code” approach mitigates tool drift, embeds access control and caching policies at the semantic boundary, and reduces the hallucination surface by restricting the agent’s action space.

The accompanying deterministic ingestion and Medallion pipeline provide the necessary foundation: telemetry is transformed into stable Gold artifacts prior to inference. The prototype implementation and case study demonstrate that this separation is operationally feasible, preserves interactive latency, and materially reduces cross-system investigative overhead. While the system is not evaluated as a large-scale benchmark against alternative grounding mechanisms, the results validate the architectural coherence of deterministic grounding in practice.

More broadly, this work argues that scalable enterprise agentic AI benefits from architectural constraints. By relocating aggregation, schema harmonization, and metric definition into explicit, version-controlled computation layers, correctness and governance shift from prompt design to system design. In regulated and operationally sensitive environments, such constraints may be preferable to fully generative query access.

\subsection{Future Directions}

Several research and engineering directions emerge from this work:

\begin{itemize}
    \item \textbf{Autonomous remediation under constraints.} Extending the architecture to support bounded corrective actions requires formal safety policies that distinguish transient anomalies from systemic faults.
    \item \textbf{Federated semantic compilation.} Multiple registry domains (e.g., infrastructure, CI/CD, security) may interoperate through federated MCP interfaces, raising questions of cross-agent trust and capability negotiation.
    \item \textbf{Causal and counterfactual reasoning.} Integrating causal models into deterministic Gold-layer transformations could support confidence estimation and counterfactual analysis in root-cause workflows.
    \item \textbf{Formal verification of interface invariants.} Static analysis or lightweight formal methods could verify properties such as tool-definition alignment and idempotent replayability in enterprise settings.
\end{itemize}

Deploying deterministic grounding in production environments also requires continued attention to privacy, access control, and compliance. Capability-scoped access and audit logging at the MCP boundary are essential to maintaining enterprise governance guarantees.

In summary, REGAL contributes a practical system architecture and design pattern for enterprise LLM grounding. Its value lies not in new learning algorithms, but in making deterministic computation and semantic compilation first-class architectural primitives for agentic systems.

\appendix
\section{Implementation Details}
\label{app:impl}

This appendix contains illustrative implementation details referenced in the main text. These examples are prototype-level and intended to clarify the architectural pattern rather than prescribe production settings.

\subsection{Tiered Parallel Fetching (illustrative)}

\begin{lstlisting}[language=Python, caption={Illustrative tiered-parallel fetching pattern (prototype)}, basicstyle=\scriptsize\ttfamily]
class FetchRawData:
    def execute(self, context):
        fetch_tasks = {
            'github': (GitHubFetcher.get_data, args),
            'jira': (JiraFetcher.get_data, args),
            'splunk': (SplunkFetcher.get_data, args)
        }

        results = {}

        # In production this should be parameterized and 
        # controlled by deployment config; the listing below 
        # is illustrative only.
        with ThreadPoolExecutor() as executor:
            future_to_source = {
                executor.submit(fn, *a): src
                for src, (fn, a) in fetch_tasks.items()
            }

            for future in as_completed(future_to_source):
                source = future_to_source[future]
                results[source] = future.result()

        return self.store_bronze(results)
\end{lstlisting}

\noindent\textbf{Note:} Worker counts and concurrency primitives should be chosen based on API rate limits, latency profiles, and the orchestration system's scheduling model. The architecture requires deterministic upserts and idempotent handlers rather than a specific concurrency mechanism.

\end{document}